\documentclass[journal]{IEEEtran}
\usepackage{algorithm}
\usepackage{moreverb}
\usepackage{amsmath}
\usepackage{colortbl}
\usepackage{wrapfig}
\usepackage{bm}
\usepackage{algorithmic}
\usepackage{tabularx}
\usepackage{cite}
\usepackage{graphicx}
\usepackage{amsfonts}
\usepackage{amsthm}
\AtBeginEnvironment{align}{\setlength{\abovedisplayskip}{4pt}}
\AtBeginEnvironment{align}{\setlength{\belowdisplayskip}{4pt}}
\usepackage{amssymb}
\usepackage{cases}
\usepackage{bbm}
\usepackage{booktabs}
\usepackage{float}
\usepackage{stfloats}

\usepackage{graphicx}
\usepackage{subcaption} 
\usepackage{caption}    

\usepackage{xcolor}
\usepackage{colortbl}
\usepackage{xpatch}
\usepackage{epstopdf}
\usepackage{algorithm}
\usepackage{algorithmic}
\usepackage{setspace}

\makeatletter
\ExplSyntaxOn
\cs_new:Npn \bibColoredItems #1#2
{
	\clist_map_inline:nn {#2} { \cs_new:cpn {bib@colored@##1} {#1} } 
}
\ExplSyntaxOff

\newcommand\bib@setcolor[1]{%
	\ifcsname bib@colored@#1\endcsname
	\expandafter\color\expandafter{\csname bib@colored@#1\endcsname}
	\else
	\normalcolor
	\fi
}

\xpatchcmd\@bibitem
{\item}
{\bib@setcolor{#1}\item}
{}{\fail}

\xpatchcmd\@lbibitem
{\item}
{\bib@setcolor{#2}\item}
{}{\fail}
\makeatother

\ifCLASSOPTIONcompsoc
\usepackage[caption=false,font=normalsize,labelfont=sf,textfont=sf]{subfig}
\else
\usepackage[caption=false,font=footnotesize]{subfig}
\fi
\ifCLASSOPTIONcaptionsoff
\usepackage[nomarkers]{endfloat}
\let\MYoriglatexcaption\caption
\renewcommand{\caption}[2][\relax]{\MYoriglatexcaption[#2]{#2}}
\fi
\usepackage{color}

\allowdisplaybreaks[4]
\begin{document}
\pagestyle{empty}
	\title{\LARGE Towards Edge Intelligence via Autonomous Navigation: \\A Robot-Assisted Data Collection Approach}
	\allowdisplaybreaks
    \author{Tingting Huang\textsuperscript{1}, Yingyang Chen\textsuperscript{1*}, Sixian Qin\textsuperscript{1}, Zhijian Lin\textsuperscript{2}, Jun Li\textsuperscript{3}, and Li Wang\textsuperscript{4}\\
    \small{\textsuperscript{1}College of Information Science and Technology, Jinan University, Guangzhou 510632, China\\
    \textsuperscript{2}College of Physics and Information Engineering, Fuzhou University, Fuzhou 350108, China\\
    \textsuperscript{3}School of Electronics and Communication Engineering, Guangzhou University, Guangzhou 510006, China\\
    \textsuperscript{4}School of Computer Science, Beijing University of Posts and Telecommunications, Beijing 100876, China
    Email:htt2024@stu2024.jnu.edu.cn; chenyy@jnu.edu.cn; a503693443@stu2022.jnu.edu.cn;\\ zlin@fzu.edu.cn; lijun52018@gzhu.edu.cn; liwang@bupt.edu.cn}
    
    \thanks{This work was supported in part by the Guangdong Basic and Applied Basic Research Foundation under Grant 2024B1515020002 and Grant 2024A1515010012; in part by the National Natural Science Foundation of China under Grant 62471140, Grant 62571149 and Grant 62525106; in part by the NSF of Fujian Province under Grant 2024J01250; in part by the State Key Laboratory of Traditional Chinese Medicine Syndrome Opening Project under Grant SKLKY2025B0003; and in part by the BUPT-China Unicom Joint Innovation Center 2025-STHZ-BJYDDX-004. \textit{(* Corresponding author: Yingyang Chen.)} }
    
		\vspace{-6mm}
    }

    \maketitle

\begin{abstract}
 
With the growing demand for large-scale and high-quality data in edge intelligence systems, mobile robots are increasingly deployed to collect data proactively, particularly in complex environments. However, existing robot-assisted data collection methods face significant challenges in achieving reliable and efficient performance, especially in non-line-of-sight (NLoS) environments. This paper proposes a communication-and-learning dual-driven (CLD) autonomous navigation scheme that incorporates region-aware propagation characteristics and a non-point-mass robot representation. This scheme enables simultaneous optimization of navigation, communication, and learning performance. An efficient algorithm based on majorization–minimization (MM) is proposed to solve the non-convex and non-smooth CLD problem. Simulation results demonstrate that the proposed scheme achieves superior performance in collision-avoidance navigation, data collection, and model training compared to benchmark methods. It is also shown that CLD can adapt to different scenarios by flexibly adjusting the weight factor among navigation, communication and learning objectives. 

\end{abstract}

\vspace{-1mm}

\vspace{-2mm}
\vspace{-1mm}
\section{Introduction}
\begin{spacing}{0.96}

Edge intelligence systems enable intelligent applications by continuously collecting large-scale and high-quality data for model training and online inference \cite{zhou2019edgeintelligence}. However, the reliability of these systems is compromised in obstacle-dense environments, particularly in non-line-of-sight (NLoS) propagation \cite{sun2015path}. The inherent constraints of sensing devices, particularly limited transmission power and communication range, further hinder stable and scalable data acquisition \cite {Luong2016data}. To overcome these issues, mobile robots have been introduced as active data-collection platforms \cite{HUANG2019mobile-robot}. By optimizing their trajectories, these robots can effectively ameliorate channel conditions and enhance data fidelity. Hence, designing efficient, reliable autonomous navigation strategies tailored to data-collection requirements is crucial in advancing edge intelligence systems.

In motion planning, traditional approaches focus primarily on collision-avoidance objectives \cite{patel2011trajectory,gerdts2012path,zhang2020optimization,han2023rda}. Patel \emph{et al.} \cite{patel2011trajectory} simplified both the robot and obstacles as point-mass models and avoided collisions by increasing the safety margin. Gerdts \emph{et al.} \cite{gerdts2012path} and Zhang \emph{et al.} \cite{zhang2020optimization} further proposed non-point-mass modeling methods in multidimensional spaces, incorporating dynamic constraints to achieve collision-free trajectory planning in complex environments. Han \emph{et al.} \cite{han2023rda} introduced the regularized dual alternating direction method of multipliers (RDA for short), which reduces computational complexity through dual decomposition and generates smooth trajectories. In communication-aware motion planning (CAMP), previous studies aimed to improve data collection efficiency \cite{saboia2022achord,wang2019backscatter,guo2021uav}. Particularly, Saboia \emph{et al.} \cite{saboia2022achord} proposed a multi-robot cooperative relaying scheme to extend communication coverage. Wang \emph{et al.} \cite{wang2019backscatter} integrated backscatter communication to optimize energy consumption. Guo \emph{et al.} \cite{guo2021uav} enhanced communication throughput through path discretization. However, these approaches often overlook collision risks. Yan \emph{et al.} \cite{yan2023communication} and Ye \emph{et al.} \cite{Ye2024ISAC} considered obstacle avoidance and communication for autonomous vehicles. Kim \emph{et al.} \cite{kim2024integrated} further investigated safe motion planning with control energy consumption considered. However, these works fail to account for the training quality of edge models, making it difficult to meet the requirements of edge intelligence.

 Albeit considerable efforts have been made,
 existing studies overlook the communication inefficiency induced by NLoS propagation\cite{patel2011trajectory,gerdts2012path,zhang2020optimization,han2023rda} and fail to consider multiple model training requirements \cite{saboia2022achord,wang2019backscatter,guo2021uav,yan2023communication,Ye2024ISAC,kim2024integrated}. To overcome these limitations, this paper proposes a novel communication and learning dual‑driven (CLD) autonomous navigation scheme for indoor data collection. {The propagation characteristics are integrated across multiple small-scale sub-regions to improve channel modeling accuracy.} By jointly optimizing motion‑and‑obstacle avoidance, communication‑aware data acquisition, and model training quality, the CLD framework enhances the comprehensive performance of edge intelligence systems. The main contributions of this paper are threefold:

 \begin{itemize}
     \item[1)] We propose a CLD autonomous navigation framework, {which incorporates region-dependent path-loss features.} The non‑point‑mass shape model is adopted to describe the geometric forms of the robot and obstacles, thereby achieving collaborative optimization between high‑precision obstacle avoidance and communication performance in autonomous navigation.

     \item[2)] We formulate a multi-objective optimization problem, which jointly optimizes navigation trajectory, communication, and learning performance with collision-free constraints. To solve this non‑convex and non-smooth problem, the majorization‑minimization (MM) algorithm and linear duality techniques are adopted.
     
     \item[3)] We validate the effectiveness of the proposed scheme through extensive comparisons. Simulation results show that CLD not only guarantees collision‑free motion planning but also improves data-collection efficiency and the quality of edge models. Moreover, the framework can adapt to new scenarios by flexibly adjusting the weights among navigation, communication, and learning.

 \end{itemize}

\end{spacing}
\vspace{-1mm}

\section{System Model}
\begin{figure}[t]
    \centering
    \scalebox{0.45}
    {\includegraphics[width=7.5in]{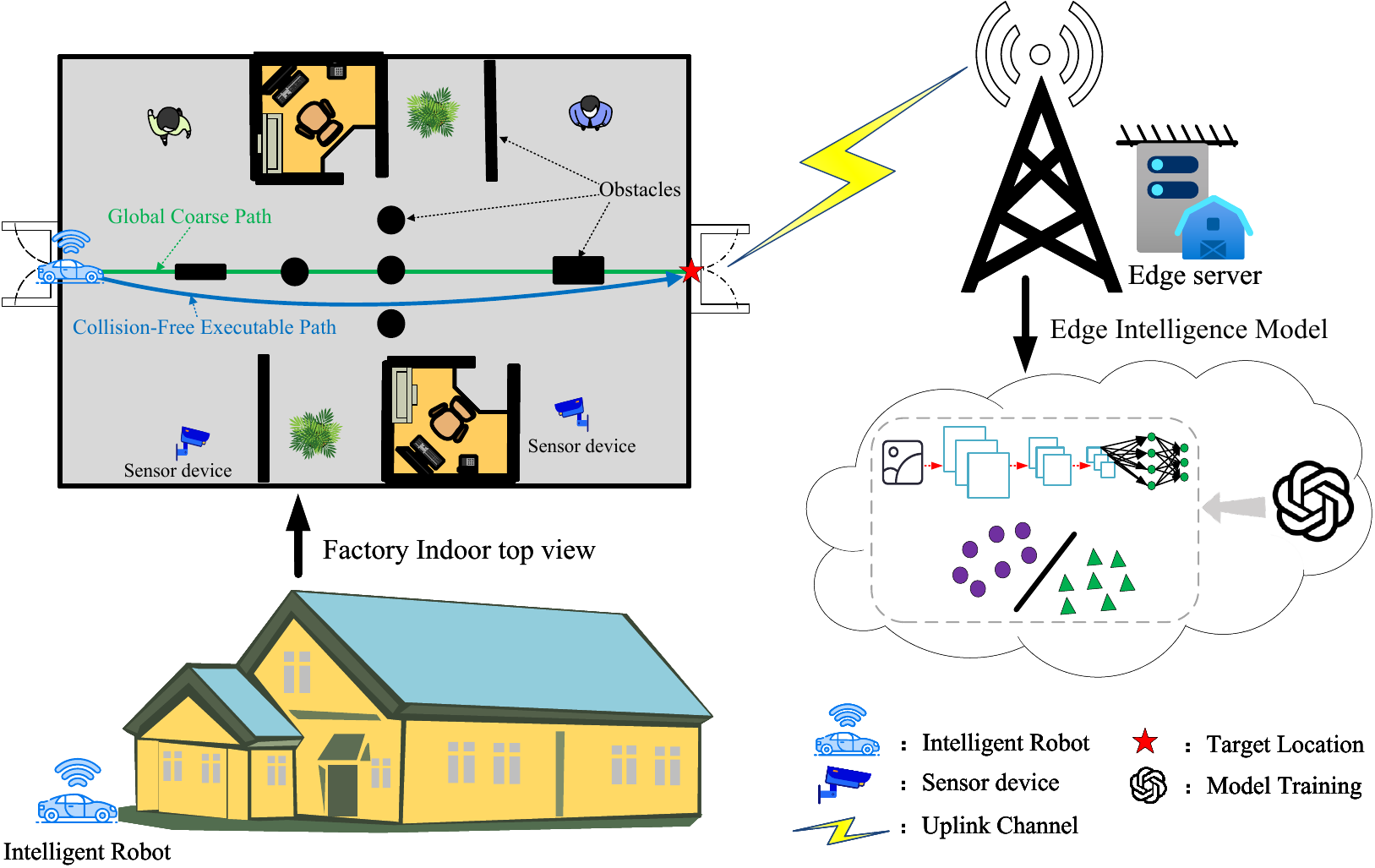}}
    \caption{An edge intelligence scenario within an industrial factory, where multiple sensor devices and obstacles exist. One intelligent robot is designed to autonomously collect distributed data and transmit it to the edge server for model training upon arrival at the target location.}
    \label{system_model}
\end{figure}

\begin{spacing}{0.965}
As shown in Fig.~\ref{system_model}, we consider an edge intelligence scenario in an industrial factory. One intelligent robot is designed to collect the disturbed data proactively and transmit it to the edge server for model training upon arrival at the target location. Assume there are $K$ sensor devices and $M$ obstacles in this scenario. The task of the robot is to generate a collision-free executable path (marked in blue) while collecting sensor data. After reaching the target location, the collected data will be transmitted via the uplink channel to an edge server (equipped with $N$ antennas) for training $E$ models. To ensure the trajectory of robot within a reasonable range, a predefined global coarse path (marked in green) is provided.

\vspace{-1mm}

\subsection{Robot Motion and Obstacle Avoidance Model}
We divide the prediction horizon into $H$ time slots, denoted as $\mathcal{H} = \{ 0,1,\dots, H-1\}$, with a continuous time step size of $T_0$. At any time step $t \in \mathcal{H}$, the robot state is represented as  $\mathbf{s}_t = ({x_t},{y_t},{\theta_t})$, where $({x_t},{y_t})$ and ${\theta_t}$ denote the robot's position and orientation, respectively. The robot's action is defined as $\mathbf{u}_t = ({v_t},{\psi_t})$, where ${v_t}$ and ${\psi_t}$ represent the linear velocity and the angular velocity, respectively. Specifically, ${{\mathbf{s}}_t} \in {\mathbb{R}^{{n_r}}}$ and ${{\mathbf{u}}_t} \in {\mathbb{R}^{{n_u}}}$, where $n_r$ depends on the dimensionality of the robot workspace and $n_u$  determined by the number of control variables for robot motion.  The navigation environment contains a set of obstacles $\mathcal{O} = \{ \mathbf{o}_1, \dots, \mathbf{o}_M \}$ and reference waypoints $\mathcal{W} = \{ \mathbf{w}_1, \mathbf{w}_2, \dots\}$. The state vector of the $m$-th ($m\in \{1,\dots, M\}$) obstacle is $\mathbf{o}_m = ({x_m},{y_m},{\theta_m})$, where $({x_m},{y_m})$ and ${\theta_m}$ represent the obstacle's position and orientation, respectively. The state vector of the $i$-th waypoint is $\smash{\mathbf{w}_i = (x_i^{\diamondsuit },y_i^{\diamondsuit },\theta _i^{\diamondsuit })}$, where $\smash{(x_i^{\diamondsuit },y_i^{\diamondsuit })}$ and $\smash{\theta _i^{\diamondsuit }}$ denote the waypoint's coordinate position and recommended orientation, respectively. The robot control follows the state evolution model ${\mathbf{s}}_{t + 1} = E({\mathbf{s}}_t, {\mathbf{u}}_t)$, where $E({\mathbf{s}}_t, {\mathbf{u}}_t)$ is determined by Ackermann kinematics
\begin{align} 
    \label{E_st_ut}
    E({\mathbf{s}}_t, {\mathbf{u}}_t) = \mathbf{A}_t \mathbf{s}_t + \mathbf{B}_t \mathbf{u}_t + \mathbf{c}_t, \quad \forall t, 
\end{align}
where the coefficient matrices $(\mathbf{A}_t,\mathbf{B}_t,\mathbf{c}_t)$ are defined in \cite[Eqs. 8-10]{han2023rda}. The robot's motion boundaries fulfill the constraints ${{\mathbf{u}}_{\min}} \leq {{\mathbf{u}}_{t+h}} \leq {{\mathbf{u}}_{\max}}$ and ${{\mathbf{a}}_{\min }} \leqslant {{\mathbf{u}}_{t + h + 1}} - {{\mathbf{u}}_{t + h}} \leqslant {{\mathbf{a}}_{\max }}{\text{  }}$ for ${\forall h\in \{0,\dots, H-1\}}$, where ${{{\mathbf{u}}_{\min }}}$ and ${{{\mathbf{u}}_{\max }}}$ denote the minimum and maximum values of the action vector, respectively, and ${\mathbf{a}}_{\min }$ and ${\mathbf{a}}_{\max }$ represent the minimum and maximum acceleration values, respectively.

To ensure the robot achieves collision-free path planning in comlpex environment, the distance between the robot and obstacles must satisfy the function $\Xi ({{\mathbf{s}}_{t }},{{\mathbf{o}}_{m}},{d_{safe}}) = {\mathbf{dist}}({{\mathbf{s}}_{t }},{{\mathbf{o}}_{m}}) - {d_{safe}}\geq 0, \forall m, t,$ where $d_{safe}$ represents the safety margin between the robot and obstacles. Traditional methods model robots and obstacles as point masses and calculate their relative positions based on center-to-center distances. The distance between the robot and the $m$-th obstacle is denoted as $\smash{\mathbf{dist}_c(\mathbf{s}_t, \mathbf{o}_{m})
= \sqrt{(x_t - x_{m})^2 + (y_t - y_{m})^2}}$. However, this distance formulation ignores the actual geometric shapes of the objects. Accordingly, we consider a non-point mass model and adopt a shape-based distance model to compute the distance. Explicitly, the occupied region of the robot at state ${\mathbf{s}}_t$ is defined as $ {\mathbb{G}_{t}}({{\mathbf{s}}_{t}},\mathbb{Z}) = \{ {\mathbf{z}} \in \mathbb{Z}|{\mathbf{R}}({{\mathbf{s}}_{t}}){\mathbf{z}} + {\mathbf{p}}({{\mathbf{s}}_{t}})\} $, where the vector $\mathbf{z}$ is the initial pose vector. The rotation matrix ${\mathbf{R}}({{\mathbf{s}}_t}) \in {\mathbb{R}^{3 \times 3}}$ and transition matrix
${\mathbf{p}}({{\mathbf{s}}_t})$ represents the orientation and position of the robot, respectively. The set $\mathbb{Z} = \{ {\mathbf{z}} \in {\mathbb{R}^3}|{\mathbf{Gz}} \leqslant {\mathbf{g}}\} $ represents the geometric shape of the robot at the initial position, where ${\mathbf{G}} \in {\mathbb{R}^{l_r \times 3}}$ and ${\mathbf{g}} \in {\mathbb{R}^{l_r}}$ denote the rotation matrices and translation vectors of all contour lines of the robot, respectively. The occupied region of the $m$-th obstacle can be expressed as the set ${\mathbb{O}_{m}} = \{ {\mathbf{z}} \in {\mathbb{R}^3}|{{\mathbf{H}}_{m,t}}{\mathbf{z}} \leqslant {{\mathbf{h}}_{m,t}}\}$, where  ${{\mathbf{H}}_{m,t}} \in {\mathbb{R}^{l_m \times 3}}$ and ${\mathbf{{{\mathbf{h}}_{m,t}}}} \in {\mathbb{R}^{l_m}}$ denote the rotation matrices and translation vectors of all contour lines of the obstacle, respectively. Specifically, $l_r$ and $l_m$ denote the number of geometric surfaces on the robot body and the obstacle, respectively. Therefore, the collision avoidance constraint can be further rewritten as follows:
\begin{align}
    \Xi ({{\mathbf{s}}_t},{{\mathbf{o}}_{m}},{d_{safe}}) = {\mathbf{dis}}{{\mathbf{t}}_s}({\mathbb{G}_{t}},{\mathbb{O}_{m}}) - {d_{safe}} \geq 0.
\end{align}

\subsection{Robot Communication}
 When the robot moves to the coordinates $({x_t},{y_t})$ in the $t$-th time slot, it triggers a sensor data collection task with duration $T_0$. During this collection period, the RF transmission module of the $k$-th ($k\in \{1,\dots,K\}$) sensor continuously transmits a symbol sequence ${{\text{z}}_{k,t}}$, with the energy constraint satisfying  $\mathbb{E}[|{z_{k,t}}{|^2}] = {p_k}$, where ${p_k}$ is the transmission power of the $k$-th sensor. The signal-to-noise ratio (SNR) at the robot receiver is given by ${\text{SN}}{{\text{R}}_k} = {{G_{t,k}}{p_k}}/{{\sigma ^2}}$, where $G_{t,k}$ represents the uplink channel gain between the $k$-th sensor and the robot's collection unit, and $\sigma^2$ denotes the power of complex Gaussian noise. Given the deployment location  $\mathbf{d}_k = ({a_k},{b_k})$  of the $k$-th sensor, the channel gain can be calculated using a distance-dependent path loss model
\begin{align}
    {G_{t,k}}({x_t},{y_t},{a_k},{b_k}) = {\beta}\left| {{x_t} - {a_k},{y_t} - {b_k}} \right|_2^{ - \alpha },
\end{align}
where $\beta$ represents the path loss parameter and $\alpha$ denotes the path loss exponent. In line-of-sight (LoS) scenarios,  $\beta_{los}=\rho _0$  with  $\alpha_{los} \in [2,5]$, where $\rho _0$ indicates the path loss at a 1-meter reference distance. In NLoS scenarios, the path loss is modeled as $\beta_{nlos} = {f_{nlos}}({x_t}, {y_t}, {a_k}, {b_k}) \times {\rho _0}$,  
$\alpha_{nlos} = { \gamma ({x_t},{y_t},{a_k},{b_k})}$, where ${ \gamma ({x_t},{y_t},{a_k},{b_k})}$ represents the path loss exponent for NLoS ($\gamma = 0$ indicates no distance dependence), and ${f_{nlos}}({x_t}, {y_t}, {a_k}, {b_k})$ is a correction factor accounting for complex radio signal propagation effects.

To accurately characterize differences in propagation characteristics between LoS and NLoS wireless channels and to adapt to channel variations in complex environments, image segmentation is employed to divide the wireless channel coverage area into $L$ small-scale sub-regions. The propagation characteristics within each sub-region are relatively consistent. Based on this, a model that integrates the features of all sub-regions is developed to describe signal propagation characteristics accurately as
\begin{align}
    \widetilde {G}_{t,k}{\text{ = }}\sum\limits_{l = 1}^L {{\beta _l^*}{\mathbb{L}_{{\mathcal{Z}_l}}}({x_t},{y_t})} \left\| {{x_t} - {a_k},{y_t} - {b_k}} \right\|_2^{ - {\alpha _l^*}},
\end{align}
where $\beta _l^*$ and $\alpha _l^*$ represent the optimal path loss parameters for $l$-th sub-region. ${\mathcal{Z}_l}$ represents the $l$-th sub-region, and ${\mathcal{Z}_l} \in \{ {\mathcal{Z}_1},{\mathcal{Z}_2},...,{\mathcal{Z}_L}\} $. The boundary of each sub-region is defined by a set of inequality constraints $ {\mathbb{Z}_l} = \{ {\mathbf{z}} \in {\mathbb{R}^2}|{{\mathbf{G}}_l}{\mathbf{z}} \leqslant {{\mathbf{g}}_l}\}$, where ${{\mathbf{G}}_l} \in {\mathbb{R}^{N_b \times 2}}$ and  ${{\mathbf{g}}_l} \in {\mathbb{R}^{N_b}}$ characterize the rotation and translation of the sub-region boundary lines, with $N_b$ denoting the number of boundaries in the sub-map. $\mathbb{L}_{\mathcal{Z}_l}(x_t, y_t)$ is the region indicator function, which equals 1 if the robot’s current position $(x_t, y_t)$ is within the $l$-th sub-region, and 0 otherwise. At $(x_t, y_t)$, the spectral efficiency of the $k$-th sensor for communication at time $t$ is 
\begin{align}
    \label{Rate_2.0}
    {R_{k,t} \left(x_t,y_t,a_k,b_k|\{ {\beta _l^*},{\alpha _l^*}\}\right)} = {\log _2}(1 + \frac{{\widetilde {G}_{t,k}{p_k}}}{{{\sigma ^2}}}).
\end{align}
For simplicity, we denote ${R_{k,t}(x_t,y_t,a_k,b_k|\{ {\beta _l^*},{\alpha _l^*}\})}$ as $\Gamma _{k,t}(x_{t}, y_{t})$.
\vspace{-1mm}
\subsection{Edge Model Training}

Upon reaching the path endpoint, the collected sensor data is transmitted to the edge server for model training. An empirical method models the classification error as a nonlinear function of the number of data samples \cite{wang2020machine}, \cite{Qin2024integrating}. Specifically, the classification error of the training model $e$ is expressed as
\begin{align}
    {\Psi_{e}} = {{a_e}{{\left( {\sum\limits_{k \in {\mathcal{G}_e}} {\sum\limits_{h = 0}^H {\frac{{B{T_0}{\Gamma  _{k,t+h}}({x_{t+h}},{y_{t+h}})}}{{{D_e}}}} }  + {A_e}} \right)}^{ - {b_e}}}},
\end{align}
where $\mathcal{G}_e$ denotes the set of sensors associated with the training of the $e$-th model, $B$ is the bandwidth allocated to each sensor group, $T_0$ is the slot duration.  $D_e$ denotes the number of bits required for each sample for model $e$, and $A_e$ denotes the historical sample count for the $e$-th model.  $a_e$ and $b_e$ are hyperparameters characterizing the learning difficulty of model $e$, with  ${a_e}>0$ and ${b_e}>0$ satisfied. We acquire the classification experimental results across various samples,
followed by predicting the values of $\left( {{a_e},{b_e}} \right)$ through a curve fitting process\cite{johnson2018predicting}.
\end{spacing}
\vspace{-1mm}

\section{Problem Formulation}
\begin{spacing}{0.96}
Based on the system model described in Section II, the robot's objective is to enable safe navigation while supporting efficient data collection and model training. Accordingly, this section formulates a multi-objective optimization problem that jointly accounts for navigation accuracy, communication efficiency, and model training performance.
In traditional model predictive control (MPC) methods, the optimization objective is typically formulated solely to minimize the deviation between the robot and the reference path \cite{zhang2020optimization}. The conventional distance cost function is defined as
\begin{align}
    C_0(\{ \mathbf{s}_{t+h} \}_{h=0}^{H}) = \sum_{h=0}^{H} \| \mathbf{s}_{t+h} - \mathbf{s}_{t+h}^{\diamondsuit} \|^2,
\end{align}
where ${ \mathbf{s}_{t+h}^{\diamondsuit} }$ represents the local reference path state extracted from the global reference path $\mathcal{W}$, determined based on the robot's current position $\mathbf{s}_t$. The cost function $C_0$ primarily supports path tracking and obstacle avoidance but does not improve the performance of data collection or model training. To maximize the volume of data collection, spectral efficiency is introduced into the objective function by involving a new term as
\begin{align}
    C_1(\{ \mathbf{s}_{t+h} \}_{h=0}^{H}) = -\sum_{h=0}^{H} \sum_{k=1}^{K} \Gamma _{k,t+h}(x_{t+h}, y_{t+h}),
\end{align}
where $\Gamma _{k,t+h}(x_{t+h}, y_{t+h})$ is the spectral efficiency of the robot when communicating with the $k$-th sensor at the position  $(x_{t+h}, y_{t+h})$  at time $t+h$. 
By minimizing $C_1$, the robot can prioritize regions that are more favorable for communication, thereby improving channel gain and data throughput.

However, given the heterogeneity of model training at edge servers, maximizing data volume purely may lead to inefficient resource utilization. Therefore, a model-training performance regularizer, $C_2$, is introduced as
\begin{align}
&{C_2}(\left\{ {{s_{t + h}}} \right\}_{h = 0}^H) \nonumber \\ 
&= \sum\limits_{e = 1}^E {\frac{{{a_e}}}{E}{{\left( {\sum\limits_{k \in {\mathcal{G}_e}} {\sum\limits_{h = 0}^H {\frac{{B{T_0}{\Gamma  _{k,t + h}}({x_{t + h}},{y_{t + h}})}}{{{D_e}}}} }  \!+\! {A_e}} \right)}^{ - {b_e}}}} \!\!, 
\end{align}
where $E$ represents the total number of AI models.
Based on the above cost functions, we formulated an optimization problem that comprehensively considers the robot's communication data acquisition, path tracking, and edge model learning performance. The optimization problem is formulated as
\begin{subequations}
    \label{P1}
    \begin{align}
    \mathcal{P}_1: \;\mathop{\min}\limits_{\{ {\mathbf{s}}_t,{\mathbf{u}}_t\} }
    &\alpha_0 C_0\bigl( \{{\mathbf{s}}_{t + h}\}_{h = 0}^H \bigr) + \alpha_1 C_1\bigl( \{{\mathbf{s}}_{t + h}\}_{h = 0}^H \bigr)\nonumber \\
    &+ \alpha_2 C_2\bigl( \{{\mathbf{s}}_{t + h}\}_{h = 0}^H \bigr) \label{P1a}\\
    \text{s.t.} \quad 
    &E({\mathbf{s}}_t, {\mathbf{u}}_t) = A_t {\mathbf{s}}_t + B_t {\mathbf{u}}_t + c_t, \quad\forall t, \label{P1b}\\
    &{\mathbf{u}}_{\min} \leqslant {\mathbf{u}}_{t + h} \leqslant {\mathbf{u}}_{\max}, \quad\forall t,\label{P1c}\\
    &{\mathbf{a}}_{\min} \leqslant {\mathbf{u}}_{t + h + 1} - {\mathbf{u}}_{t + h} \leqslant {\mathbf{a}}_{\max},\quad\forall t, \label{P1d}\\
    &\Xi ({\mathbf{s}}_{t },{\mathbf{o}}_{m},d_{safe}) \geqslant 0, \quad\forall t,m \in \mathcal{O}\label{P1e},
\end{align}
\end{subequations}
where $ \alpha_0 > 0, \alpha_1 > 0, \alpha_2  > 0$ are hyperparameters used to maintain the weights among~$ C_0, C_1$ and $C_2$, supporting task-oriented parameter tuning. Particularly, \eqref{P1b} is the robot state evolution model constraint, and \eqref{P1e} represents the collision avoidance constraint.
\end{spacing}
\vspace{-1mm}
\section{Cross-Layer Optimization Algorithm}
\begin{spacing}{0.965} 
Due to the existence of regularization terms $C_1$ and $C_2$ in the objective function and the collision avoidance constraint \eqref{P1e}, the problem $\mathcal{P}_1$ is typically non-convex. Furthermore, constraint \eqref{P1b} contains nonlinear terms, while constraints \eqref{P1c} and \eqref{P1d} involve linear inequalities. To address this non-convex problem, this paper employs the MM algorithm to convexify the objective function first and then utilizes linear duality techniques to handle nonlinear constraints in autonomous navigation.

\subsection{Constraint Convexification via Dual Theory}
To capture the spatiotemporal distribution of collision risk while balancing local obstacle avoidance and global path flexibility, an $L_1$-norm regularization term is incorporated into the objective function. Leveraging the sparsity-inducing property of $L_1$ regularization, the safety distance parameters $\{d_{t+h}\}$ exhibit a nonuniform distribution over the time horizon, thereby adaptively emphasizing collision avoidance at near-term steps. The specific implementation is to add the following penalty function to the objective function of $\mathcal{P}_1$ 
\begin{align}
    {C_3}(\{ {d_{t + h}}\} _{h = 0}^H) =  -\sum\limits_{h = 0}^H {{\left| {{d_{t+h}}} \right|}},
\end{align}
where ${\{d_{t + h}}\} _{h = 0}^H$ represents the time-varying safety distance sequence constructed within the prediction horizon. This sequence satisfies the boundary constraint $d_{\min}\leqslant d_{t+h}\leqslant d_{\max}$, where $d_{\max}$ and $d_{\min}$ are the maximum and minimum values of the safety distance, respectively. After introducing this time-varying safety distance mechanism, the collision avoidance constraint \eqref{P1e} of the problem $\mathcal{P}_1$ can be reformulated as follows
\begin{subequations}
\begin{align}
\label{P1e-2.0a}
&\mathbf{dist}_s(\mathbb{G}_t, \mathbb{O}_{m}) - d_t \geqslant 0, \quad \forall m,t, \\
\label{P1e-2.0b}
&d_{\min}\leqslant d_t \leqslant d_{\max}, \quad \forall m,t
\end{align}
\end{subequations}

To address the non-smoothness introduced by constraint \eqref{P1e-2.0a}, it is equivalently transformed into a smooth linear dual form. This facilitates the subsequent parallel computation process \cite{li2024edge}, and its specific expression is given as follows
\begin{subequations}
    \begin{align}
      & {\bm{\lambda} _{m,t}} \geqslant 0,{\bm{\mu} _{m,t}} \geqslant 0,{z_{m,t}} \geqslant 0,\quad \forall m,t \label{eq:a} \\ 
      & \bm{\lambda} _{m,t}^T{{\mathbf{H}}_{m,t}}{\mathbf{p}}({{\mathbf{s}}_t})\! -\! \bm{\lambda} _{m,t}^T{{\mathbf{h}}_{m,t}}\! -\! \bm{\mu} _{m,t}^T{\mathbf{g}}\! -\! {z_{m,t}}\! =\! {d_{t}},   \forall m,t\label{eq:b} \\ 
      & \bm{\mu} _{m,t}^T{\mathbf{G}} + \bm{\lambda} _{m,t}^T{{\mathbf{H}}_{m,t}}{\mathbf{R}}({{\mathbf{s}}_t}) = 0,\quad\forall m,t \label{eq:c} \\ 
      & \left\| {{\mathbf{H}}_{m,t}^T{\bm{\lambda} _{m,t}}} \right\| \leqslant 1,\quad\forall m,t \label{eq:d} 
    \end{align}
\end{subequations}
where ${\bm{\lambda} _{m,t}} \in {\mathbb{R}^{{l_m}}}$ and ${\bm{\mu} _{m,t}} \in {\mathbb{R}^{{l_r}}}$ are dual variables introduced by the geometric constraints of the robot body and obstacle surfaces, respectively. ${z_{m,t}} \in \mathbb{R}$ is a slack variable introduced in \eqref{eq:b} to convert inequality constraints into equality constraints.

\subsection{MM-Based Non-Convex Problem Solving}
To address the non-convexity of $C_1$, we construct surrogate functions $\{ {{\tilde \Gamma  }_{k,t}}\} $ to replace $\{ {\Gamma  _{k,t}}\} $, thereby obtaining a sequence of surrogate problems. Specifically, given any feasible solution $\{ {\mathbf{s}}_{k,t + h}^*,{\mathbf{u}}_{k,t + h}^*\} $ of $\mathcal{P}1$, the surrogate function is defined as
\begin{align}
&\tilde{\Gamma}_{k,t}({\mathbf{s}_t}|\mathbf{s}_t^*) \nonumber\\
& = \log_2\left(1+ \sum\limits_{l=1}^L\beta_l\mathbb{L}_{Z_l}(x_t,y_t)p_k \times \sigma^{-2} \times \Big[2\,\|\mathbf{s}_t^*-\mathbf{c}_{k,t}\|_2^{-\alpha_l}\right. \nonumber \\
&\quad \left.-\|\mathbf{s}_t^*-\mathbf{c}_{k,t}\|_2^{-2\alpha_l}\,\|x_t-a_k,y_t-b_k\|_2^{\alpha_l}\Big]\right),
\end{align}
where ${{\mathbf{c}}_{k,t}} = {[{x_k},{y_k},\theta _t^*]^T}$. The surrogate function $\{ {{\tilde \Gamma  }_{k,t}}\} $ must satisfy the following constraints:
\begin{itemize}
    \item[(1)] Lower bound constraint: ${{\tilde \Gamma  }_{k,t}}({{\mathbf{s}}_t}|{\mathbf{s}}_t^*) \leqslant {\Gamma  _{k,t}}({{\mathbf{s}}_t})$;
    \item[(2)] Concavity constraint: ${{\tilde \Gamma  }_{k,t}}({{\mathbf{s}}_t}|{\mathbf{s}}_t^*)$ is concave with respect to ${{\mathbf{s}}_t}$;
    \item[(3)] Local consistency constraint: At the optimal solution,
${{\tilde \Gamma  }_{k,t}}({\mathbf{s}}_t^*|{\mathbf{s}}_t^*) = {\Gamma  _{k,t}}({\mathbf{s}}_t^*)$ and $\nabla {{\tilde \Gamma  }_{k,t}}({\mathbf{s}}_t^*|{\mathbf{s}}_t^*) = \nabla {\Gamma  _{k,t}}({\mathbf{s}}_t^*)$.
\end{itemize}
The lower bound property is verified using the inequality $\frac{1}{x} \geqslant \frac{1}{y} - \frac{1}{{{y^2}}}(x - y)$ for $\forall x, y >0$, which guarantees that ${{\tilde \Gamma  }_{k,t}}({{\mathbf{s}}_t}|{\mathbf{s}}_t^*) \leqslant {\Gamma  _{k,t}}({{\mathbf{s}}_t})$. Concavity is established by analyzing the Hessian matrix of ${{\tilde \Gamma  }_{k,t}}$, which is shown to be negative semidefinite. By verifying that both the function values and gradients of ${{\tilde \Gamma  }_{k,t}}$ and ${\Gamma  _{k,t}}$ are equal at ${\mathbf{s}}_t^*$,  it is confirmed that the surrogate function maintains consistency with the original function at the optimal solution, thereby satisfying the local consistency condition.
Based on the lower-bound property of the surrogate function, replacing the original functions $\{ {{\tilde \Gamma  }_m}\} $ with the surrogate functions $\{ {\Gamma  _m}\} $ in the vicinity of a feasible point yields a lower-bound estimate of the objective function. By iteratively constructing new surrogate functions starting from the current solution, the lower bound is progressively tightened until the convergence condition $\|\Gamma_{k,t} - \tilde{\Gamma}_{k,t}\|_2 < \varepsilon$ is satisfied. Therefore, assuming the solution at the $n$-th iteration is $\{ {\mathbf{s}}_{k,t + h}^{[n]},{\mathbf{u}}_{k,t + h}^{[n]}\} $, the original problem $\mathcal{P}1$ can be transformed into $\mathcal{P}2$ in the $(n+1)$-th iteration as follows
\begin{subequations}
\begin{align*}
\mathcal{P}_2: &\mathop{\min}\limits_{\substack{\{ {\mathbf{s}}_t,{\mathbf{u}}_t\} \\ \{ {\bm{\lambda}_{m,t}},{\bm{\mu}_{m,t}},{z_{m,t}}\} }} 
  \alpha_0 C_0 - \alpha_1\sum\limits_{h = 0}^H \sum\limits_{k = 1}^K \tilde{\Gamma}_{k,t + h}({\mathbf{s}}_{t + h}|{\mathbf{s}}_{t + h}^{[n]}) \nonumber \\
&\ + \alpha_2\sum\limits_{e = 1}^E \frac{a_e}{E}\left( \sum\limits_{k \in \mathcal{G}_e} \sum\limits_{h = 0}^H \frac{B{T_0} \tilde{\Gamma}_{k,t + h}({\mathbf{s}}_{t + h}|{\mathbf{s}}_{t + h}^{[n]})}{D_e} + A_e \right)^{-b_e} \nonumber \\
&\ - \alpha_3 C_3\left( \{{d_{t + h}}\}_{h = 0}^H \right) \\
\text{s.t.}\ &\eqref{P1b}, \eqref{P1c}, \eqref{P1d}, \eqref{P1e-2.0b}, \eqref{eq:a}, \eqref{eq:b}, \eqref{eq:c}, \eqref{eq:d}. 
\end{align*}
\end{subequations}
where~$\alpha_3$~is a hyperparameter for adjusting the weight of the collision avoidance capability under different scenarios. Now, problem $\mathcal{P}_2$ is smooth and convex, which can be solved directly using the CVXPY numerical optimization toolkit. The optimal solution is denoted as $\{ {\mathbf{s}}_{t + h}^*,{\mathbf{u}}_{t + h}^*\} $, and the iterative solution is updated as $\{ {\mathbf{s}}_{t + h}^{[n + 1]} = {\mathbf{s}}_{t + h}^*,{\mathbf{u}}_{t + h}^{[n + 1]} = {\mathbf{u}}_{t + h}^*\} $. This procedure is repeated to solve problem $\mathcal{P}2$ until convergence is achieved. According to the local consistency constraint, as long as the initial point $\{ {\mathbf{s}}_{t + h}^{[0]},{\mathbf{u}}_{t + h}^{[0]}\}$ is feasible for $\mathcal{P}_2$, every solution in the iterative sequence $(\{ {\mathbf{s}}_{t + h}^{[0]},{\mathbf{u}}_{t + h}^{[0]}\},\{ {\mathbf{s}}_{t + h}^{[1]},{\mathbf{u}}_{t + h}^{[1]}\},...,\{ {\mathbf{s}}_{t + h}^{[n]},{\mathbf{u}}_{t + h}^{[n]}\} )$ satisfies the Karush–Kuhn–Tucker (KKT) conditions of $\mathcal{P}_2$.

In the following, we analyze the complexity of the overall algorithm. Specifically, in the $(n+1)$-th iteration, problem $\mathcal{P}2$ involves $Hn_r$  state variables, $Hn_u$ control variables, and $H$ distance variables. In addition, the optimization problem includes $H \sum\nolimits_{m = 1}^M l_m$ robot shape variables and $H l_r$ obstacle shape variables, which arise from the geometric modeling of the robot body and surrounding obstacles. Consequently, the total computational complexity of $\mathcal{P}2$ can be bounded by $\mathcal{O}(\mathcal{I}({(H({{n}_r} + {n_u} + 1) + H(\sum\nolimits_{m = 1}^M {{l_m}}  + {l_r}))^{3.5}}))$ \cite{ben2001lectures}, where $\mathcal{I}$ denotes the number of iterations required for the algorithm convergence. 

\end{spacing}
\vspace{-1mm}

\section{Simulation Results}
\begin{spacing}{0.965}
In this section, we evaluate the performance of the proposed scheme. Specifically, there are $E=2$ models and $K=2$ sensors. In particular, we consider two classification models: a CNN and an SVM. Each model is assigned to a specific sensor. The specific parameters for each classification error function
as follows: $\left( {{a_1},{b_1}} \right) = \left( {1.718,0.3781} \right)$ for CNN, $\left( {{a_2},{b_2}} \right) = \left( {4.55,0.7268} \right)$ for SVM.

\begin{figure*}[t]
    \centering
    \begin{subfigure}[t]{0.32\textwidth}
        \centering
        \includegraphics[width=\linewidth]{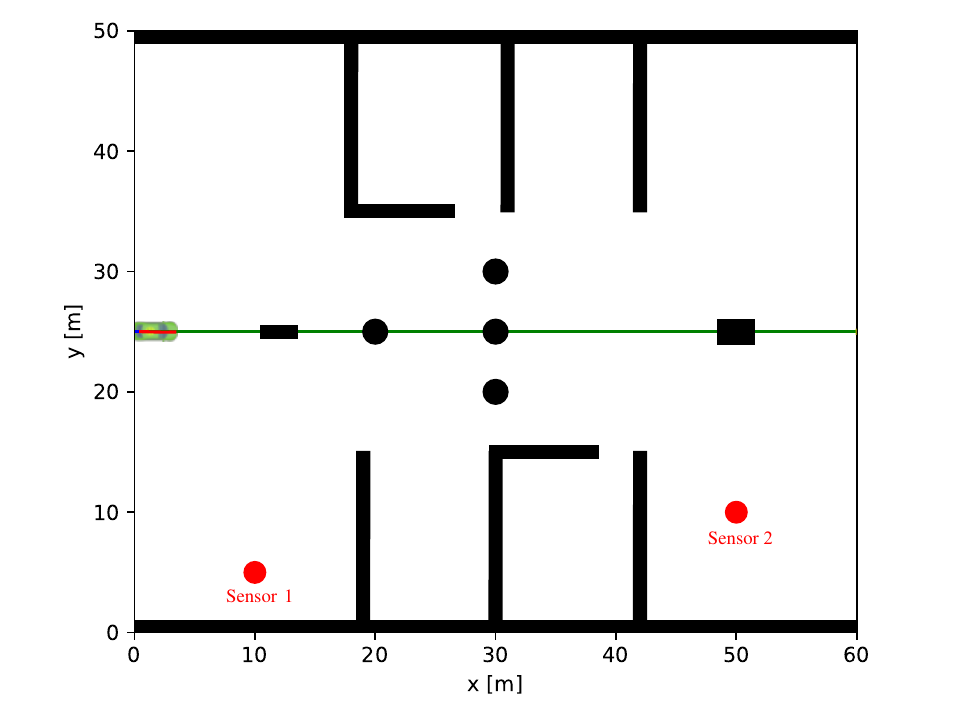}
        \caption{Scene graph}
        \label{subfig:a}
    \end{subfigure}
    \hfill
    \begin{subfigure}[t]{0.32\textwidth}
        \centering
        \includegraphics[width=\linewidth]{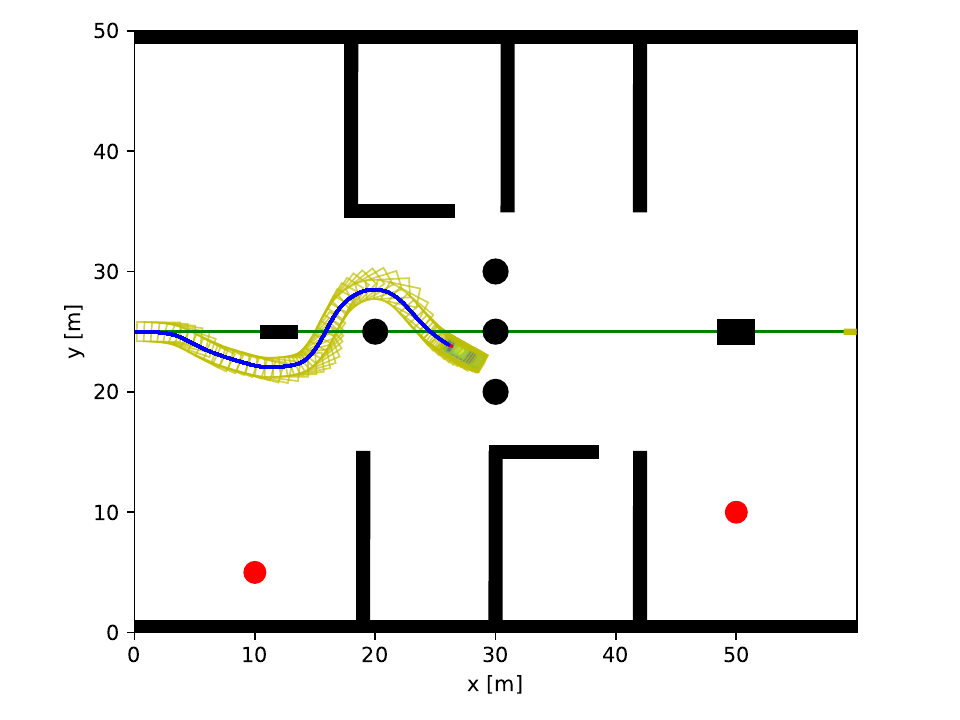}
        \caption{PMM+Commu}
        \label{subfig:pmm_com}
    \end{subfigure}
    \hfill
    \begin{subfigure}[t]{0.31\textwidth}
        \centering
        \includegraphics[width=\linewidth]{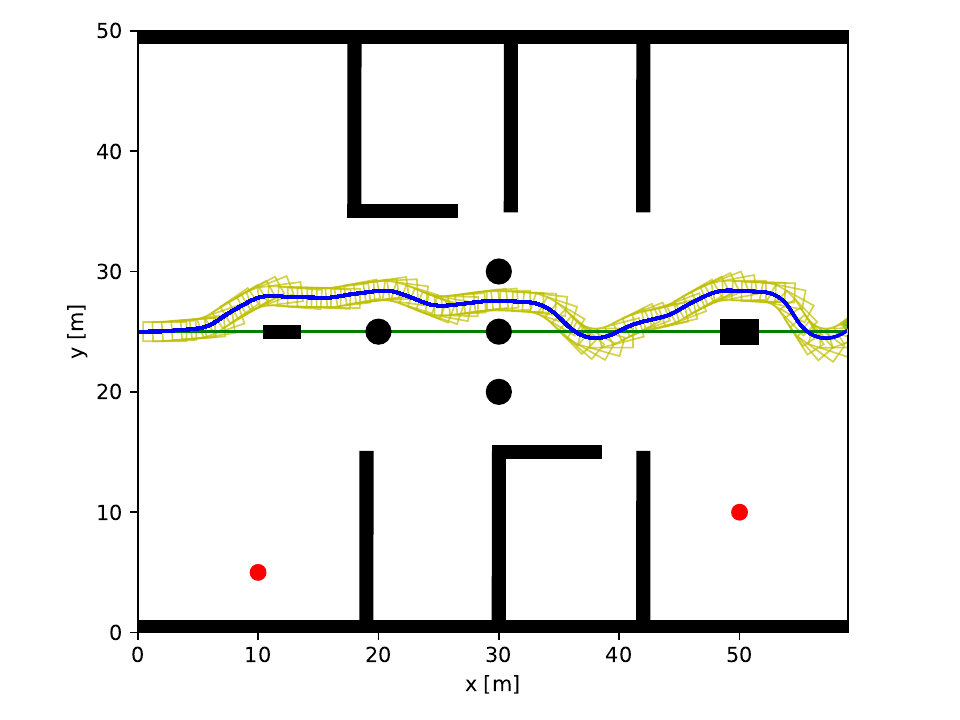}
        \caption{OBCA}
        \label{subfig:OBCA}
    \end{subfigure}


    \begin{subfigure}[t]{0.32\textwidth}
        \centering
        \includegraphics[width=\linewidth]{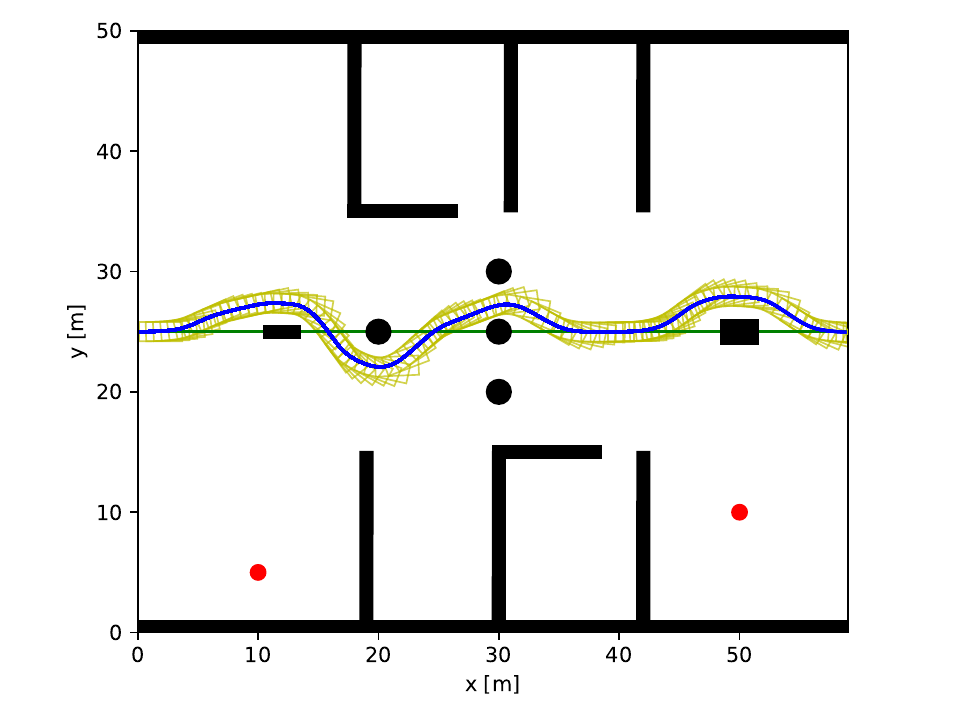}
        \caption{RDA}
        \label{subfig:RDA}
    \end{subfigure}
    \hfill
    \begin{subfigure}[t]{0.32\textwidth}
        \centering
        \includegraphics[width=\linewidth]{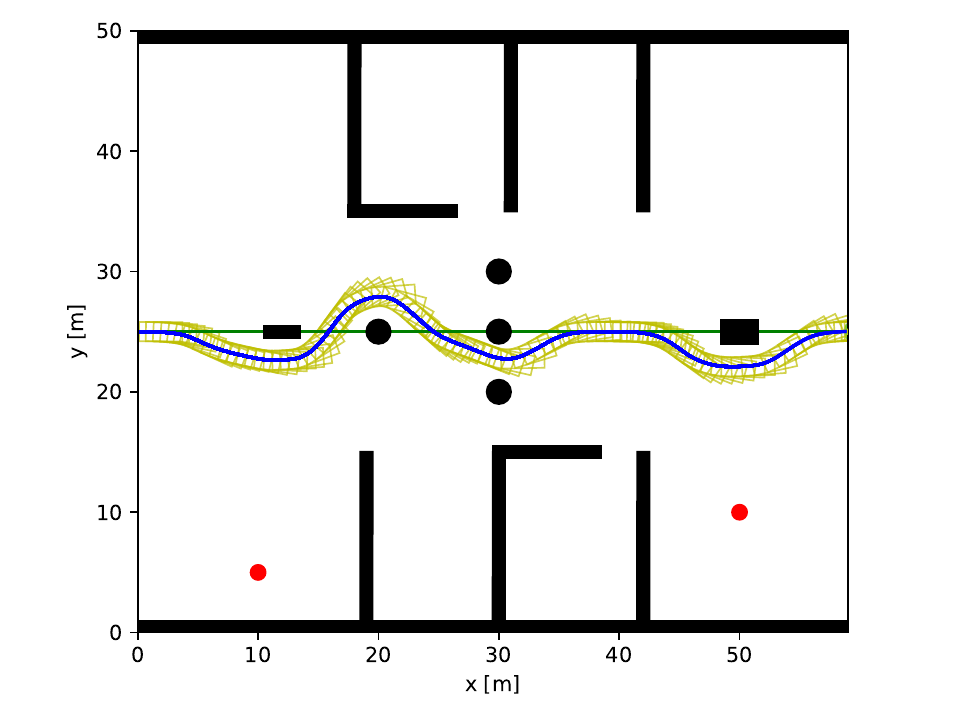}
        \caption{RDA+Commu}
        \label{subfig:RDA_com}
    \end{subfigure}
    \hfill
    \begin{subfigure}[t]{0.32\textwidth}
        \centering
        \includegraphics[width=\linewidth]{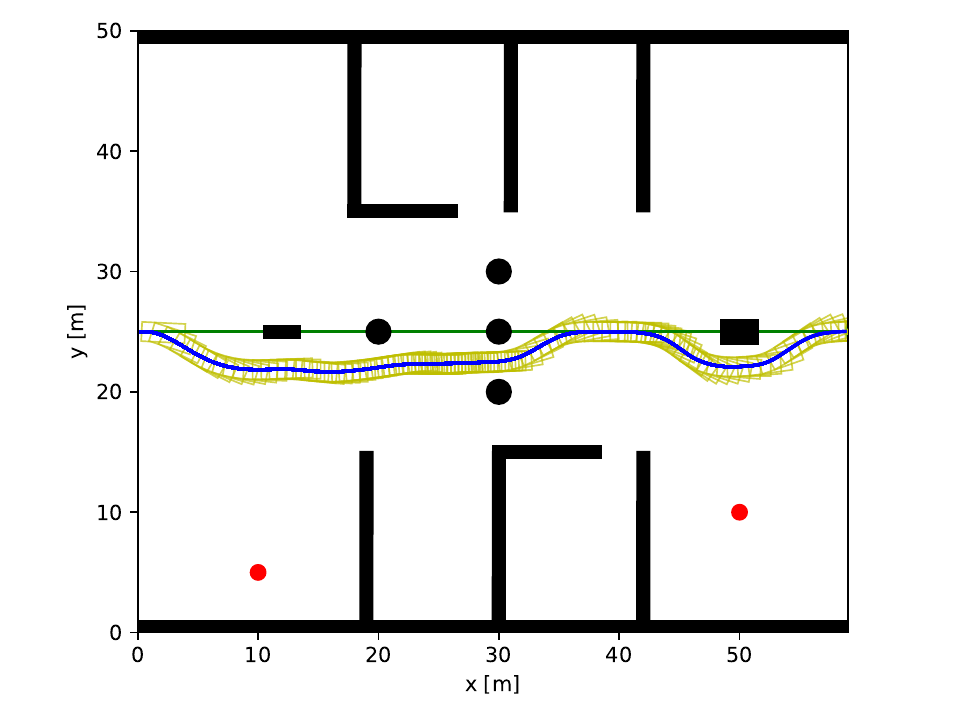}
        \caption{CLD (proposed)}
        \label{subfig:CLD}
    \end{subfigure}

    \caption{Autonomous navigation trajectories under different schemes.}
    \label{Fig_Schemes}
    \vspace{-4mm}
\end{figure*}

The edge server is assumed to have $N=16$ antennas. Each sensor transmits at a power of $P = 0.02$ W; the noise variance is $N_0 = -70$ dBm;   the duration of each slot
is $T_0 = 0.2$ s; and the bandwidth allocated to each device is $B = 0.1$ MHz. For model training, the number of pre-trained historical samples is set to $A_1 = 100$ and $A_2 = 100$ for the CNN and SVM models, respectively. All experiments are conducted on a desktop computer with an Intel Core i7-12700 CPU, using the PyCharm 2022 development environment and Python 3.9.12. Model training is performed on an NVIDIA GeForce GTX 1650 SUPER GPU. Fig.~\ref{subfig:a} depicts the simulation scenario. We compare the proposed CLD scheme with the following baselines
\begin{itemize}
    \item PMM+Commu, which simplifies both obstacles and the robot as point masses to reduce collision avoidance modeling complexity \cite{licea2019communication}; 
    \item OBCA, which employs optimization strategies for safe navigation and exploits a traditional collision avoidance method based on model predictive control\cite{zhang2020optimization}; 
    \item RDA, which models obstacles as convex sets and achieves high-precision collision avoidance through accurate computation of minimum inter-object distances\cite{han2023rda}; 
    \item RDA+Commu, which extends RDA by incorporating communication awareness.
\end{itemize}

Fig. \ref {Fig_Schemes} displays the autonomous navigation performance under different schemes. In particular, Fig.~\ref{subfig:a} depicts the scene graph, where the sensor locations are shown in red and the obstacles are marked in black with various shapes. The green straight line represents the global reference path, yellow boxes indicate the robot body, and blue curves depict the robot’s actual trajectory. Fig.~\ref{subfig:pmm_com}, it is evident that the PMM+Commu scheme leads the robot into a dead loop in obstacle-dense areas due to the simplified point models and large safety distances. For the OBCA scheme in Fig.~\ref{subfig:OBCA}, it failed to closely follow the reference path, and the motion trajectory deviated significantly from the sensor area. In Fig.~\ref{subfig:RDA}, though the RDA scheme produces a smooth and robust trajectory, it fails to prioritize the paths near sensors as much as possible. Compared to RDA, the RDA+commu scheme in Fig.~\ref{subfig:RDA_com} selects a path closer to the sensor, but its path planning still prioritizes proximity to the reference path. By contrast, the proposed CLD scheme in Fig.~\ref{subfig:CLD} successfully avoids obstacles and reaches the target position by prioritizing paths near Sensors 1 and 2. It improves channel gain and data throughput, thereby enhancing communication performance.

\begin{figure}[t]
	\centering
	\scalebox{0.45}
	{\includegraphics[width=7.2in]{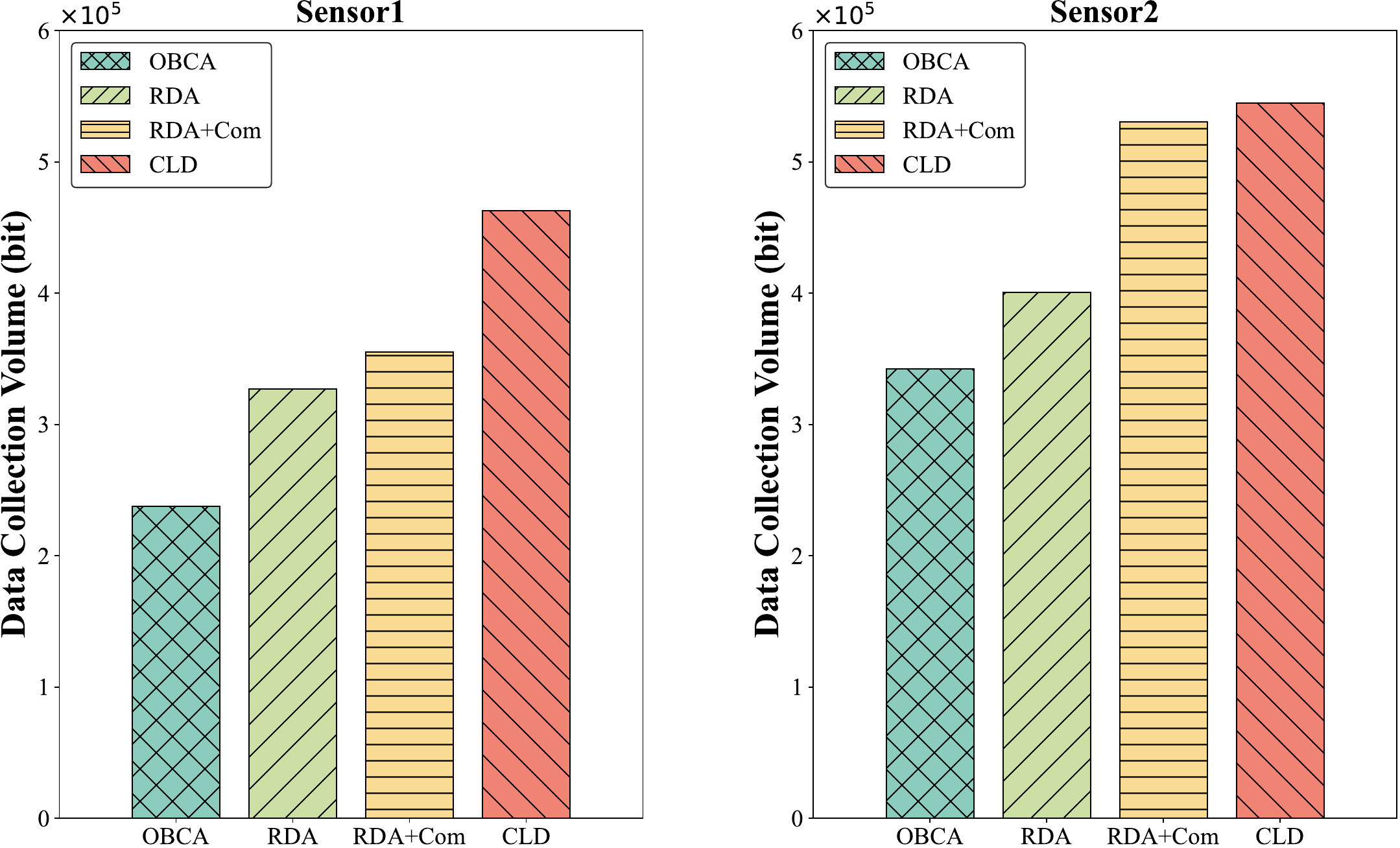}}
    \vspace{-1mm}
	\caption{The data amount collected by the robot under different schemes}
	\label{Fig_ct}
	\vspace{-2mm}
\end{figure}

Fig.~\ref{Fig_ct} compares the amount of data collected across different schemes. It is evident that the proposed scheme achieves the highest amount at both sensors compared to other baselines. Compared to the OBCA scheme, the CLD intelligently plans a path that tends toward high-signal regions, avoiding signal attenuation from obstacles and maximizing data collection volume. The RDA scheme prioritizes obstacle avoidance and path tracking as its main optimization objectives, with data collection serving merely as an indirect benefit. In contrast, the CLD scheme offers more flexible and efficient path selection, resulting in superior data-collection performance. Though RDA+Commu explicitly considers data collection requirements, it fails to adequately address AI training performance. Driven by the learning performance regularizer, CLD prioritizes data that provides greater training benefits for the CNN model, further enhancing the effectiveness of data collection. 

\begin{figure}[t]
	\centering
	\scalebox{0.45}
	{\includegraphics[width=7.2in]{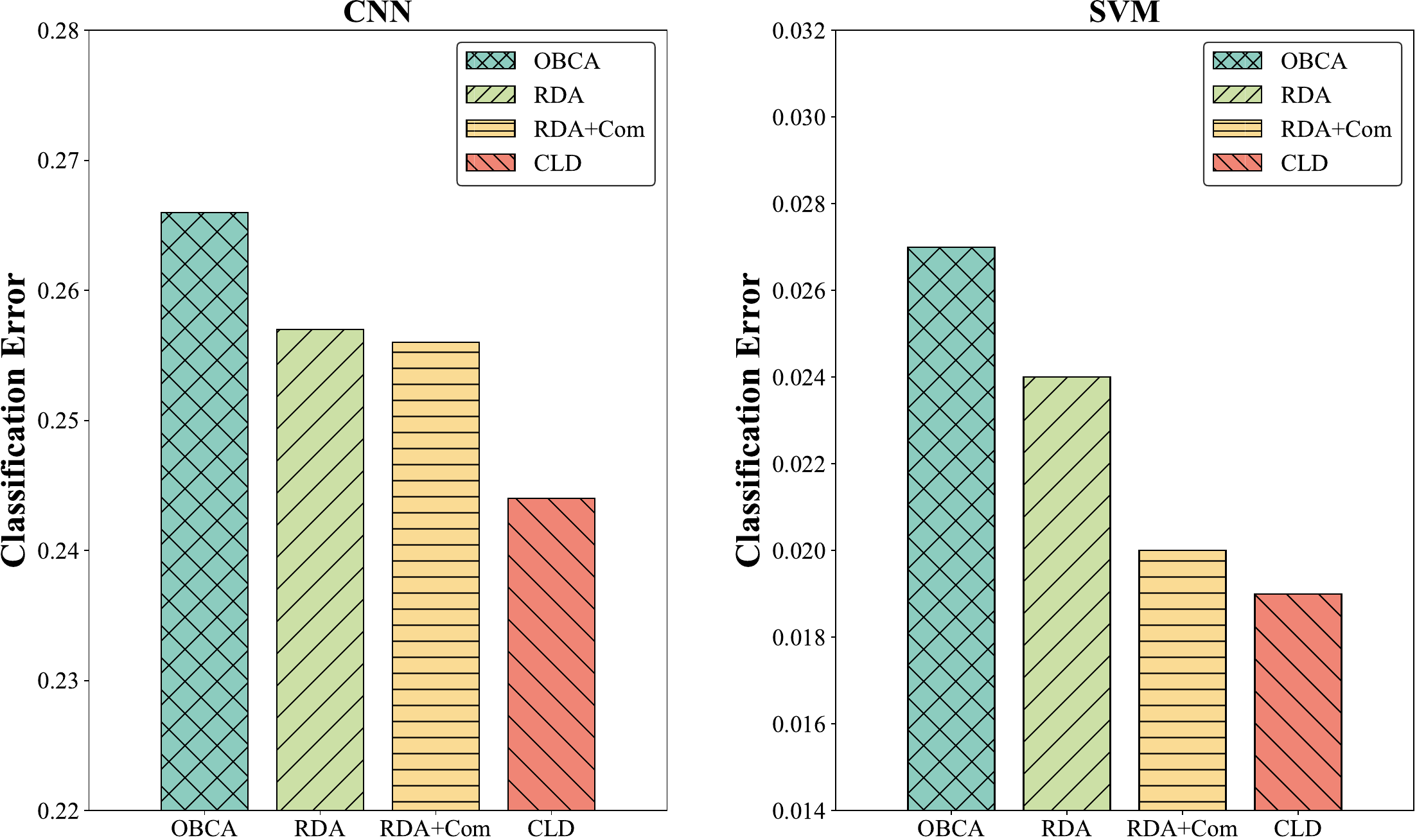}}
	\caption{Model classification errors under different schemes}
    \vspace{-2mm}
	\label{Fig_ce}
\end{figure}

Fig.~\ref{Fig_ce} compares the classification errors across different schemes. 
For both models, the CLD scheme achieves significantly low classification errors, demonstrating the superiority in terms of learning performance. The OBCA scheme, which focuses solely on obstacle avoidance without adequately accounting for communication and data-collection requirements, yields limited data, thereby constraining model learning performance. Though the RDA scheme improves computational efficiency and generates paths that closely follow the reference trajectory, it fails to incorporate learning performance into its optimization objectives, thus underutilizing the potential of joint communication-learning performance. The RDA+Commu scheme improves data-collection efficiency but does not sufficiently account for data quality or model-training requirements. In contrast, the proposed CLD scheme not only optimizes communication throughput during path planning, but also can flexibly balance between communication and learning by fine-tuning the weights. 

\end{spacing}

\vspace{-1mm}
\section{Conclusion}
\begin{spacing}{0.965}
In this paper, we have proposed a CLD autonomous navigation scheme for edge intelligence systems. A multi-objective optimization problem has been formulated to jointly consider navigation accuracy, communication efficiency, and learning performance. Simulation results have shown that the proposed scheme effectively improves the communication throughput and model learning performance while ensuring collision-free navigation. Moreover, the proposed scheme has demonstrated the capability to dynamically balance navigation, communication, and learning objectives under different task requirements, indicating its adaptability to diverse application scenarios.
\end{spacing}

	%
	%

	\ifCLASSOPTIONcaptionsoff
	
	\fi
	\bibliographystyle{IEEEtran}
     \begin{spacing}{0.95}
	\bibliography{reference.bib}
    \end{spacing}

	%
	%
	%
	%
	%
	
	
	


\end{document}